\documentstyle[acl,psfig]{article}
\newcounter{examplectr}
\newcounter{subexamplectr}
\newenvironment{ex}%
   {\vspace{.1in}\addtocounter{examplectr}{1}
     \setcounter{subexamplectr}{0}
     \begin{list}
       {(\arabic{examplectr})}%
       {\setlength{\topsep}{0in}
        \setlength{\leftmargin}{.25in}
               \setlength{\labelsep}{0.075in}}
       \item \begin{minipage}[t]{6.4cm} 
   }%
   {\end{minipage}
    \end{list}\vspace{.1in}}
\newenvironment{subex}%
   { \addtocounter{subexamplectr}{1}
     \begin{list}
       {\alph{subexamplectr}}%
       {\setlength{\topsep}{-\parskip}
        \setlength{\leftmargin}{0.175in}
        \setlength{\labelsep}{0.075in}}
       \item
   }%
   {\end{list}}
\newcommand{\exnum}[2]{\addtocounter{examplectr}{#1}(\arabic{examplectr}{#2})\addtocounter{examplectr}{-#1}}
\newcommand {\etal} {{\it et al.}}
\title{\LARGE\bf APPORTIONING DEVELOPMENT EFFORT\\[1mm]
IN A PROBABILISTIC LR PARSING SYSTEM\\[0.9mm]
THROUGH EVALUATION\\[1mm]
}
\author{John Carroll \\ Cognitive and Computing Sciences \\
University of Sussex \\ Brighton BN1 9QH, UK \\[0.5mm]
{\it john.carroll@cogs.susx.ac.uk}
\And Ted Briscoe \\ Computer Laboratory \\ University of Cambridge \\
Pembroke Street, Cambridge CB2 3QG, UK \\[0.5mm]
{\it ejb@cl.cam.ac.uk}}
\begin{document}
\maketitle
\begin{abstract}
We describe an implemented system for robust domain-independent syntactic
parsing of English, using a unification-based grammar of part-of-speech and
punctuation labels coupled with a probabilistic LR parser. We present
evaluations of the system's performance along several different dimensions;
these enable us to assess the contribution that each individual part
is making to the success of the system as a whole, and thus prioritise the effort
to be devoted to its further enhancement. Currently, the system is able to parse
around 80\% of sentences in a substantial corpus of general text containing a
number of distinct genres. On a random sample of 250 such sentences the
system has a mean crossing bracket rate of 0.71 and recall and precision of 83\%
and 84\% respectively when evaluated against manually-disambiguated
analyses\footnote{Some of this work was carried out while the second author was
visiting Rank Xerox, Grenoble. The work was also supported by UK DTI/SALT
project 41/5808 `Integrated Language Database', and by SERC/EPSRC Advanced
Fellowships to both authors. Geoff Nunberg provided encouragement and much
advice on the analysis of punctuation, and Greg Grefenstette undertook the
original corpus tokenisation and segmentation for the punctuation experiments.
Bernie Jones and Kiku Ribas made helpful comments on an earlier draft. We are
responsible for any mistakes.}.
\end{abstract}

\section{1.\ INTRODUCTION}

This work is part of an effort to develop a robust, domain-independent
syntactic parser capable of yielding the unique correct analysis for
unrestricted naturally-occurring input. Our goal is to develop a
system with performance comparable to extant part-of-speech taggers,
returning a syntactic analysis from which predicate-argument structure
can be recovered, and which can support semantic interpretation. The
requirement for a domain-independent analyser favours statistical
techniques to resolve ambiguities, whilst the latter goal favours a
more sophisticated grammatical formalism than is typical in
statistical approaches to robust analysis of corpus material.

Briscoe \& Carroll (1993) describe a probablistic parser using a
wide-coverage uni\-fication-based grammar of English written in the
Alvey Natural Language Tools (ANLT) metagrammatical formalism (Briscoe
\etal, 1987), generating around 800 rules in a syntactic variant of
the Definite Clause Grammar formalism (DCG, Pereira \& Warren, 1980)
extended with iterative (Kleene) operators. The ANLT grammar is linked
to a lexicon containing about 64K entries for 40K lexemes, including
detailed subcategorisation information appropriate for the grammar,
built semi-automatically from a learners' dictionary (Carroll \&
Grover, 1989).  The resulting parser is efficient,
constructing a parse forest in roughly quadratic time (empirically), and
efficiently returning the ranked {\it n}-most likely analyses (Carroll, 1993,
1994).  The probabilistic model is a refinement of probabilistic
context-free grammar (PCFG) conditioning CF `backbone' rule
application on LR state and lookahead item. Unification of the
`residue' of features not incorporated into the backbone is performed
at parse time in conjunction with reduce operations. Unification
failure results in the associated derivation being assigned a
probability of zero. Probabilities are assigned to transitions in the
LALR(1) action table via a process of supervised training based on
computing the frequency with which transitions are traversed in a
corpus of parse histories. The result is a probabilistic parser which,
unlike a PCFG, is capable of probabilistically discriminating
derivations which differ only in terms of order of application of the
same set of CF backbone rules, due to the parse context defined by the
LR table.

Experiments with this system revealed three major problems which our
current research is addressing.  Firstly, improvements in
probabilistic parse selection will require a `lexicalised'
grammar/parser in which (minimally) probabilities are associated with
alternative subcategorisation possibilities of individual lexical
items. Currently, the relative frequency of subcategorisation
possibilities for individual lexical items is not recorded in
wide-coverage lexicons, such as ANLT or COMLEX (Grishman \etal, 1994).
Secondly, removal of punctuation from the input (after segmentation
into text sentences) worsens performance as punctuation both reduces
syntactic ambiguity (Jones, 1994) and signals non-syntactic
(discourse) relations between text units (Nunberg, 1990). Thirdly, the
largest source of error on unseen input is the omission of appropriate
subcategorisation values for lexical items (mostly verbs), preventing
the system from finding the correct analysis. The current coverage---the
proportion of sentences for which at least one analysis was
found\footnote{Briscoe \& Carroll (1995) note that ``coverage'' is a weak measure
since discovery of one or more global analyses does not entail that the correct
analysis is recovered.}---of this system on a general corpus (e.g.\ Brown or LOB)
is estimated to be around 20\% by Briscoe (1994). Therefore, we have developed a
variant probabilistic LR parser which does not rely on subcategorisation and
uses punctuation to reduce ambiguity.  The analyses produced by this parser can
be utilised for phrase-finding applications, recovery of subcategorisation
frames, and other `intermediate' level parsing problems.

\section{2.\ PART-OF-SPEECH TAG SEQUENCE GRAMMAR}

We utilised the ANLT metagrammatical formalism to develop a
feature-based, declarative description of part-of-speech (PoS) label
sequences (see e.g.\ Church, 1988) for English. This grammar compiles into a
DCG-like grammar of approximately 400 rules. It has been designed to enumerate
possible valencies for predicates (verbs, adjectives and nouns) by including
separate rules for each pattern of possible complementation in
English. The distinction between arguments and adjuncts is expressed,
following X-bar theory (e.g.\ Jackendoff, 1977), by Chomsky-adjunction
of adjuncts to maximal projections (\mbox{XP $\rightarrow$ XP
Adjunct}) as opposed to government of arguments (i.e.\ arguments are
sisters within X1 projections; \mbox{X1 $\rightarrow$ X0 Arg1$\ldots$
ArgN}).  Although the grammar enumerates complementation
possibilities and checks for global sentential well-formedness, it is
best described as `intermediate' as it does not attempt to associate
`displaced' constituents with their canonical position / grammatical
role. 

The other difference between this grammar and a more conventional one
is that it incorporates some rules specifically designed to overcome
limitations or idiosyncrasies of the tagging process. For example,
past participles functioning adjectivally, as in \exnum{+1}{a}, are
frequently tagged as past participles (VVN) as in \exnum{+1}{b}, so
the grammar incorporates a rule (violating X-bar theory) which parses
past participles as adjectival premodifiers in this context.
\begin{ex}
\begin{subex}
The disembodied head
\end{subex}
\begin{subex}
The\_AT disembodied\_VVN head\_NN1
\end{subex}
\end{ex}
Similar idiosyncratic rules are incorporated for dealing with gerunds,
adjective-noun conversions, idiom sequences, and so forth. 
Further details of the PoS grammar are given in Briscoe \&
Carroll (1994, 1995).

The grammar currently covers around 80\% of the Susanne
corpus (Sampson, 1995), a 138K word treebanked and balanced subset of
the Brown corpus. Many of the `failures' are due to the root
S(entence) requirement enforced by the parser when dealing with fragments from
dialogue and so forth. We have not relaxed this requirement since it increases
ambiguity, our primary interest at this point being the extraction of
subcategorisation information from full clauses in corpus data.

\section{3.\ TEXT GRAMMAR AND PUNCTUATION}

Nunberg (1990) develops a partial `text' grammar for English
which incorporates many constraints that (ultimately) restrict
syntactic and semantic interpretation. For example, textual adjunct
clauses introduced by colons scope over following punctuation, as
\exnum{+1}{a} illustrates; whilst textual adjuncts introduced by
dashes cannot intervene between a bracketed adjunct and the textual
unit to which it attaches, as in \exnum{+1}{b}.
\begin{ex}
\begin{subex}
*He told them his reason: he would not renegotiate his contract,  but
he did not explain to the team owners. (vs. but would stay)
\end{subex}
\begin{subex}
*She left --  who could blame her -- (during the chainsaw scene) and went home.
\end{subex}
\end{ex}

We have developed a declarative grammar in the ANLT metagrammatical
formalism, based on Nunberg's procedural description.  This grammar
captures the bulk of the text-sentential constraints described by
Nunberg with a grammar which compiles into 26 DCG-like
rules. Text
grammar analyses are useful because they demarcate some of the
syntactic boundaries in the text sentence and thus reduce ambiguity,
and because they identify the units for which a syntactic analysis
should, in principle, be found; for example, in \exnum{+1}{}, the
absence of dashes would mislead a parser into seeking a syntactic
relationship between {\it three} and the following names, whilst in
fact there is only a discourse relation of elaboration between this
text adjunct and pronominal {\it three}.
\begin{ex} 
The three -- Miles J. Cooperman, Sheldon
Teller, and Richard Austin -- and eight other defendants were charged
in six indictments with conspiracy to violate federal narcotic law.
\end{ex}

Further details of the text grammar are given in Briscoe \& Carroll
(1994, 1995). The text grammar has been tested on the Susanne corpus and covers
99.8\% of sentences. (The failures are mostly text segmentation
problems). The number of analyses varies from one (71\%) to the
thousands (0.1\%).  Just over 50\% of Susanne sentences contain some
punctuation, so around 20\% of the singleton parses are
punctuated. The major source of ambiguity in the analysis of
punctuation concerns the function of commas and their relative scope
as a result of a decision to distinguish delimiters and separators
(Nunberg 1990:36). Therefore, a text sentence containing eight commas
(and no other punctuation) will have 3170 analyses. The multiple uses
of commas cannot be resolved without access to (at least) the
syntactic context of occurrence.

\section{4.\ THE INTEGRATED GRAMMAR}

Despite Nunberg's observation that text grammar is distinct from
syntax, text grammatical ambiguity favours interleaved application of
text grammatical and syntactic constraints. Integrating the text
and the PoS sequence grammars is straightforward and the result remains modular,
in that the text grammar is `folded into' the PoS sequence grammar, by
treating text and syntactic categories as overlapping and dealing with
the properties of each using disjoint sets of features, principles of
feature propagation, and so forth. In addition to the core
text-grammatical rules which carry over unchanged from the stand-alone text
grammar, 44 syntactic rules (of pre- and post- posing, and
coordination) now include (often optional) comma markers corresponding
to the purely `syntactic' uses of punctuation.

The approach to text grammar taken here is in many ways similar to
that of Jones (1994). However, he opts to treat punctuation marks as
clitics on words which introduce additional featural information into
standard syntactic rules. Thus, his grammar is thoroughly integrated
and it would be harder to extract an independent text grammar or build
a modular semantics. Our less-tightly integrated grammar is described in more
detail in Briscoe \& Carroll (1994). 

\section{5.\ PARSING THE SUSANNE AND SEC CORPORA}

We have used the integrated grammar to parse the Susanne corpus and the quite
distinct Spoken English Corpus (SEC; Taylor \& Knowles, 1988), a 50K word
treebanked corpus of transcribed British radio programmes punctuated by the
corpus compilers. Both corpora were retagged using the Acquilex HMM tagger
(Elworthy, 1993, 1994) trained on text tagged with a slightly modified version
of CLAWS-II labels (Garside \etal, 1987). In contrast to previous
systems taking as input fully-determinate sequences of PoS
labels, such as Fidditch (Hindle, 1989) and MITFP (de Marcken, 1990),
for each word
the tagger returns multiple label hypotheses, and each is thresholded before
being passed on to the parser: a given label is retained if it is the
highest-ranked, or, if the highest-ranked label is assigned a likelihood of less
than 0.9, if its likelihood is within a factor of 50 of this. We thus attempt to
minimise the effect of incorrect tagging on the parsing component by allowing
label ambiguities, but control the increase in indeterminacy and concomitant
decrease in subsequent processing efficiency by applying the thresholding
technique. On Susanne, retagging allowing only a single label per word results
in a 97.90\% label/word assignment accuracy, whereas multi-label tagging with
this thresholding scheme results in 99.51\% accuracy.

In an earlier paper (Briscoe \& Carroll, 1995) we gave results for a
previous version of the grammar and parsing system. We have made a number of
significant improvements to the system since then, the most fundamental being
the use of multiple labels for each word. System accuracy evaluation results are
also improved since we now output trees that conform more closely to the
annotation conventions employed in the test treebank.

\subsection{COVERAGE AND AMBIGUITY}

To examine the efficiency and coverage of the grammar we applied it to
our retagged versions of Susanne and SEC. We used the ANLT chart
parser (Carroll, 1993), but modified just to count the number of
possible parses in the parse forests (Billot \& Lang, 1989) rather
than actually unpacking them. We also imposed a per-sentence time-out
of 30 seconds CPU time, running in Franz Allegro Common Lisp 4.2 on an
HP PA-RISC 715/100 workstation with 128 Mbytes of physical memory.

\begin{table*}[tb]
\centering
\begin{tabular}{|l|rr|rr|} \hline
                   & Susanne       && SEC & \\ \hline

Parse fails        & 1476 & 21.0\%  &  809 & 31.3\% \\

1--9 parses        & 1436 & 20.5\%  &  477 & 18.4\% \\

10--99 parses      & 1218 & 17.4\%  &  378 & 14.6\% \\

100--999 parses    &  953 & 13.6\%  &  276 & 10.7\% \\

1K--9.9K parses    &  694 &  9.9\%  &  225 &  8.7\% \\

10K--99K parses    &  474 &  6.8\%  &  154 &  6.0\% \\

100K+ parses       &  750 & 10.7\%  &  264 & 10.2\% \\

Time-outs          &   13 &  0.2\%  &    4 &  0.2\% \\
\hline 
Number of sentences        & 7014  &&  2717 & \\

Mean sentence length (MSL) & 20.1  &&  22.6 & \\

MSL -- fails               & 20.9  &&  29.5 & \\

MSL -- time-outs           & 73.6  &&  65.8 & \\

Average Parse Base        & 1.313  && 1.300 & \\ \hline
\end{tabular}
\caption{Grammar coverage on Susanne and SEC}
\label{sus-sec}
\end{table*}

For both corpora, the majority of
sentences analysed successfully received under 100 parses, although there is a
long tail in the distribution.  Monitoring this distribution is helpful during
grammar development to ensure that coverage is increasing but the
ambiguity rate is not. A more succinct though less intuitive measure
of ambiguity rate for a given corpus is Briscoe \& Carroll's (1995)
average parse base (APB), defined as the geometric mean over all sentences in
the corpus of $\sqrt[n]{p}$, where $n$ is the number of words in a sentence, and
$p$, the number of parses for that sentence. Thus, given a sentence $n$ words
long, the APB raised to the $n$th power gives the number of analyses
that the grammar can be expected to assign to a sentence of
that length in the corpus. Table~\ref{sus-sec} gives these measures for all of
the sentences in Susanne and in SEC.

As the grammar was developed solely with reference to Susanne,
coverage of SEC is quite robust. The two corpora differ considerably
since the former is drawn from American written text whilst the latter
represents British transcribed spoken material. The corpora overall contain
material drawn from widely disparate genres / registers, and are more
complex than those used in DARPA ATIS tests, and more diverse than those used
in MUCs and probably also the Penn Treebank. Black \etal\ (1993)
report a coverage of around 95\% on computer manuals, as opposed to our
coverage rate of 70--80\% on much more heterogeneous data and longer sentences.
The APBs for Susanne and SEC of 1.313
and 1.300 respectively indicate that sentences of average
length in each corpus could be expected to be assigned of the order of 238 and
376 analyses (i.e.\ $1.313^{20.1}$ and $1.300^{22.6}$). 

The parser throughput on these tests, for sentences successfully
analysed, is around 25 words per CPU second on an HP PA-RISC
715/100. Sentences of up to 30 tokens (words plus sentence-internal
punctuation) are parsed in an average of under 1 second each, whilst those around
60 tokens take on average around 7 seconds. Nevertheless, the relationship
between sentence length and processing time is fitted well by a
quadratic function, supporting the findings of Carroll (1994) that in
practice NL grammars do not evince worst-case parsing complexity.

\subsection{Grammar Development \& Refinement}

The results we report above relate to the latest version of the tag
sequence grammar. To date, we have spent about one person-year
on grammar development, with the effort spread fairly evenly over a
two-and-a-half-year period. The various phases in the development and
refinement of the grammar can be observed in an analysis of the coverage and APB
for Susanne and SEC over this period---see table~\ref{cov-amb}.
The phases, with dates, were:
\begin{description}
\item[6/92--11/93] Initial development of the grammar.
\item[11/93--7/94] Substantial increase in coverage on the development corpus
(Susanne), corresponding to a drive to increase the general coverage of the
grammar by analysing parse failures on actual corpus material. From a lower
initial figure, coverage of SEC (unseen corpus), increased by a larger
factor. 
\item[7/94--12/94] Incremental improvements in coverage, but at the cost of
increasing the ambiguity of the grammar.
\item[12/94--10/95] Improving the accuracy of the system by trying to ensure
that the correct analysis was in the set returned.
\end{description}

\begin{table}[tb]
\begin{center}
\begin{tabular}{|l|rr|r|} \hline
         & \multicolumn{2}{c|}{Susanne} & \multicolumn{1}{c|}{SEC} \\
date     & coverage  & $APB^{20.1}$  & coverage \\ \hline

11/93    & 47.8\%    & 667        & 34.3\% \\

1/94     & 56.7\%    & 160        & 45.7\% \\

7/94     & 75.3\%    & 192        & 67.1\% \\

12/94    & 79.0\%    & 217        & 68.9\% \\

10/95    & 79.0\%    & 238        & 68.7\% \\ \hline
\end{tabular}
\caption{Grammar coverage and ambiguity during development}
\label{cov-amb}
\end{center}
\end{table}

Since the coverage on SEC is increasing at the same time as on Susanne,
we can conclude that the grammar has not been specifically tuned to
the particular sublanguages or genres represented in the development corpus.
Also, although the almost-50\% initial coverage on the heterogeneous text of
Susanne compares well with the state-of-the-art in grammar-based
approaches to NL analysis (e.g.\ see Taylor \etal, 1989; Alshawi \etal, 1992),
it is clear that the
subsequent grammar refinement phases have led to major improvements in
coverage and reductions in spurious ambiguity.

We have experimented with increasing the richness of the lexical feature
set by incorporating subcategorisation information for verbs into the
grammar and lexicon. We constructed randomly from Susanne a test corpus of
250 in-coverage sentences, and in this, for each word tagged as possibly
being an open-class verb (i.e.\ not a modal or auxiliary) we extracted from the
ANLT lexicon (Carroll \& Grover, 1989) all verbal entries for that word. We
then mapped these entries into our PoS grammar experimental
subcategorisation scheme, in which we distinguished each possible pattern
of complementation allowed by the grammar (but not control
relationships, specification of prepositional heads of PP complements etc.\
as in the full ANLT representation scheme). We then attempted to parse the
test sentences, using the derived verbal entries instead of the
original generic entries which generalised over all the subcategorisation
possibilities. 31 sentences now failed to receive a parse, a decrease in
coverage of 12\%. This is due to the fact that the ANLT lexicon, although
large and comprehensive by current standards (Briscoe \& Carroll, 1996),
nevertheless contains many errors of omission.

\subsection{PARSE SELECTION}

A probabilistic LR parser was trained with the integrated grammar by
exploiting the Susanne treebank bracketing. An LR parser (Briscoe \&
Carroll, 1993) was applied to unlabelled bracketed sentences from the
Susanne treebank, and a new treebank of 1758 correct and complete analyses
with respect to the integrated grammar was constructed semi-automatically by
manually resolving the remaining ambiguities. 250 sentences from the new
treebank, selected randomly, were kept back for testing\footnote{The
appendix contains a random sample of sentences from the test corpus.}. The
remainder, together with a further set of analyses from 2285 treebank sentences
that were not checked manually, were used to train a probabilistic version of
the LR parser, using  Good-Turing smoothing to estimate the probability of unseen
transitions in the LALR(1) table (Briscoe \& Carroll, 1993; Carroll, 1993).  The
probabilistic parser can then return a ranking of all possible analyses for a
sentence, or efficiently return just the {\it n}-most probable (Carroll,
1993).

The probabilistic parser was tested on the 250 sentences held out from the
manually-disambiguated treebank (of lengths 3--56 tokens, mean 18.2). The
parser was set up to return only the highest-ranked analysis for each sentence.
Table~\ref{sus-eval}
\begin{table*}[tb]
\centering
\begin{tabular}{|l|rrrr|} \hline
       & \multicolumn{1}{c}{Zero} & \multicolumn{1}{c}{Mean} &
\multicolumn{1}{c}{Recall} & \multicolumn{1}{c|}{Precision} \\

       & \multicolumn{1}{c}{crossings} & \multicolumn{1}{c}{crossings} &&
\\ \hline

{\em Probabilistic parser analyses} &&&& \\
Top-ranked analysis
       & 59.6\%        & 1.03          & 74.0\%         & 73.0\% \\
Random analysis
       & 40.4\%        & 1.84          & 58.6\%         & 60.0\% \\
&&&& \\ \hline

{\em Manually-disambiguated analyses} &&&& \\
`Ideal' analysis
       & 80.1\%        & 0.41          & 85.4\%         & 82.9\% \\ \hline
\end{tabular}
\caption{GEIG evaluation metrics for test set of 250 held-back sentences
against Susanne bracketings}
\label{sus-eval}
\end{table*}
shows the results of this test---with respect to the
original Susanne bracketings---using the Grammar Evaluation Interest Group
scheme (GEIG, see e.g.\ Harrison \etal, 1991)\footnote{We would like to thank
Phil Harrison for supplying the evaluation software.}. This compares unlabelled
bracketings derived from corpus treebanks with those derived from parses for the
same sentences by computing {\it recall}, the ratio of matched brackets over all
brackets in the treebank; {\it precision}, the ratio of matched brackets over
all brackets found by the parser; {\it mean crossings}, the number of times a
bracketed sequence output by the parser overlaps with one from the treebank but
neither is properly contained in the other, averaged over all sentences; and
{\it zero crossings}, the percentage of sentences for which the analysis
returned has zero crossings.

The table also gives an indication of the best and worst possible
performance of the disambiguation component of the system, showing the
results obtained when parse selection is replaced by a simple random
choice, and the results of evaluating the analyses in the manually-disambiguated
treebank against the corresponding original Susanne bracketings. In this latter
figure, the mean number of crossings (0.41) is greater than zero mainly because
of incompatibilities between the structural representations chosen by the
grammarian and the corresponding ones in the treebank. Precision is less than
100\% due to crossings, minor mismatches and inconsistencies (due to the manual
nature of the markup process) in tree annotations, and the fact that
Susanne often favours a ``flat'' treatment of VP constituents, whereas our
grammar always makes an explicit choice between argument- and adjunct-hood.
Thus, perhaps a more informative test of the accuracy of our probabilistic
system would be evaluation against the manually-disambiguated corpus of analyses
assigned by the grammar. In this, the mean crossing figure drops to 0.71 and the
recall and precision rise to 83--84\%, as shown in table~\ref{disambig-eval}.  

\begin{table*}[tb]
\centering
\begin{tabular}{|l|rrrr|} \hline
       & \multicolumn{1}{c}{Zero} & \multicolumn{1}{c}{Mean} &
\multicolumn{1}{c}{Recall} & \multicolumn{1}{c|}{Precision} \\

       & \multicolumn{1}{c}{crossings} & \multicolumn{1}{c}{crossings} &&
\\ \hline

{\em Probabilistic parser analyses} &&&& \\
Top-ranked analysis
       & 67.2\%        & 0.71          & 82.9\%         & 83.9\% \\ \hline
\end{tabular}
\caption{GEIG evaluation metrics for test set of 250 held-back sentences
against the manually-disambigated analyses}
\label{disambig-eval}
\end{table*}

Black \etal\ (1993:7) use the crossing brackets measure to define a notion
of structural consistency, where the structural consistency rate for
the grammar is defined as the proportion of sentences for which at
least one analysis---from the many typically returned by the grammar---contains
no crossing brackets, and report a rate of around 95\% for the IBM grammar
tested on the computer manual corpus. However, a problem with the GEIG
scheme and with structural consistency is that both are still weak measures
(designed to avoid problems of parser/treebank representational compatibility)
which lead to unintuitive numbers whose significance still depends heavily on
details of the relationship between the representations compared
(e.g.\ between structure assigned by a grammar and that in a treebank).
One particular problem with the crossing bracket measure is that a single
attachment mistake embedded $n$ levels deep (and
perhaps completely innocuous, such as an ``aside'' delimited by dashes) can lead
to $n$ crossings being assigned, whereas incorrect identification of arguments
and adjuncts can go unpunished in some cases. 

Schabes \etal\ (1993) and Magerman (1995) report results using the
GEIG evaluation scheme which are numerically similar in terms of parse
selection to those reported here, but achieve 100\% coverage. However, their
experiments are not strictly comparable because they both utilise more
homogeneous and probably simpler corpora. (The appendix gives an indication of
the diversity of the sentences in our corpus). In addition, Schabes
\etal\ do not recover tree labelling, whilst Magerman has developed a parser
designed to produce identical analyses to those used in the Penn Treebank,
removing the problem of spurious errors due to grammatical
incompatibility.  Both these approaches achieve better coverage by
constructing the grammar fully automatically, but as an inevitable
side-effect the range of text phenomena that can be parsed becomes limited
to those present in the training material, and being able to deal with new
ones would entail further substantial treebanking efforts.

To date, no robust parser has been shown to be practical and useful for
some NLP task. However, it seems likely that, say, rule-to-rule semantic
interpretation will be easier with hand-constructed grammars with an
explicit, determinate rule-set.  A more meaningful parser comparison would
require application of different parsers to an identical and extended
test suite and utilisation of a more stringent standard evaluation
procedure sensitive to node labellings.

\subsection{Training Data Size and Accuracy}

Statistical HMM-based part-of-speech taggers require of the order of 100K words
and upwards of training data (Weischedel \etal, 1993:363); taggers
inducing non-probabilistic rules (e.g.\ Brill, 1994) require similar
amounts (Gaizauskas, pc). Our probabilistic disambiguation system
currently makes no use of lexical frequency information, training only
on structural configurations. Nevertheless, the number of parameters in the
probabilistic model is large: it is the total number of possible
transitions in an LALR(1) table containing over 150000 actions. It is
therefore interesting to investigate whether the system requires more
or less training data than a tagger.

\begin{figure*}[tb]
\centering
\leavevmode
\psfig{file=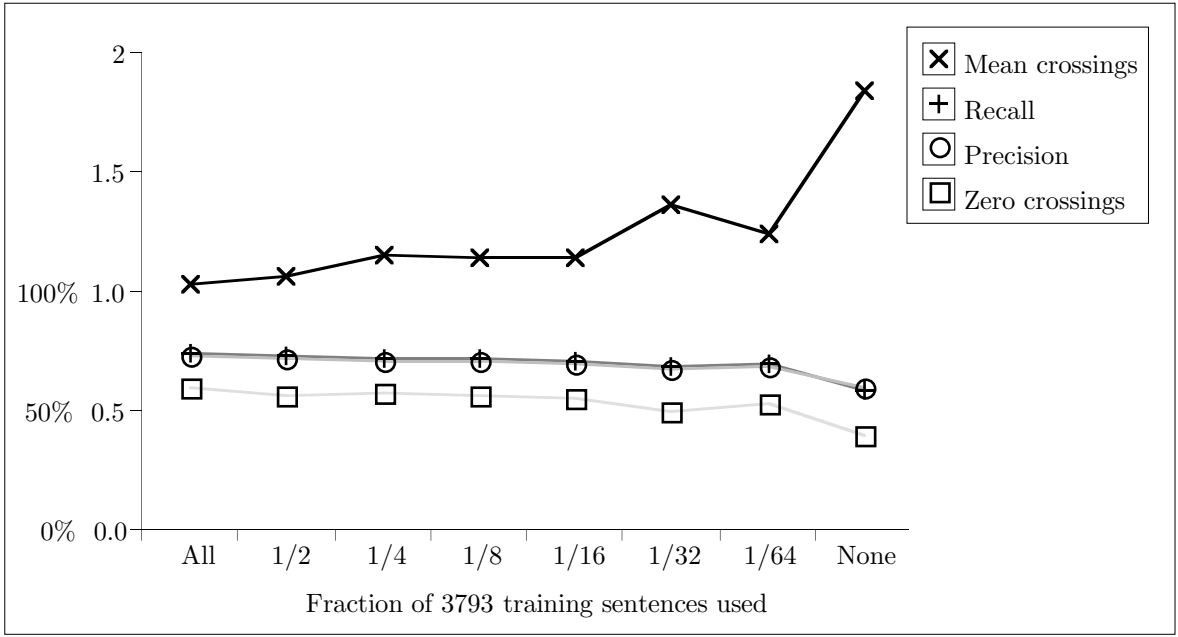}
\caption{GEIG metrics for held-back sentences, training on varying amounts
of data}
\label{train-eval}
\end{figure*}

We therefore ran the same experiment as above, using GEIG to measure the
accuracy of the system on the 250 held-back sentences, but varying the
amount of training data with which the system was provided. We started at
the full amount (3793 trees), and then successively halved it by selecting
the appropriate number of trees at random. The results obtained are
given in figure~\ref{train-eval}.

The results show convincingly that the system is extremely robust when
confronted with limited amounts of training data: when using a mere one
sixty-fourth of the full amount (59 trees), accuracy was degraded by only
10--20\%. However, there is a large decrease in accuracy with no training data
(i.e.\ random choice). Conversely, accuracy is still improving at 3800 trees,
with no sign of over-training, although it appears to be approaching an upper
asymptote. To determine what this might be, we ran the system on a set of
250 sentences randomly extracted from the {\it training} corpus. On this set,
the system achieves a zero crossings rate of 60.0\%, mean crossings 0.88, and
recall and precision of 77.0\% and 75.2\% respectively, with respect to the
original Susanne bracketings. Although this is a different set of sentences, it
is likely that the upper asymptote for accuracy for the test corpus lies in this
region. Given that accuracy is increasing only slowly and is relatively
close to the asymptote it is therefore unlikely that it would be worth
investing effort in increasing the size of the training corpus at this stage in
the development of the system.

\section{6.\ CONCLUSIONS}

In this paper we have outlined an approach to robust domain-independent
parsing, in which subcategorisation
constraints play no part, resulting in coverage that greatly improves upon more
conventional grammar-based approaches to NL text analysis. We described an
implemented system, and evaluated its performance along several different
dimensions. We assessed its coverage and that of previous versions on a
development corpus and an unseen corpus, and demonstrated that the grammar
refinement we have carried out has led to substantial improvements in coverage
and reductions in spurious ambiguity. We also evaluated the accuracy of
parse selection with respect to treebank analyses, and, by varying the amount of
training material, we showed that it requires comparatively little data to
achieve a good level of accuracy.

We have made good progress in increasing grammar coverage, though we
have now reached a point of diminishing returns. Further significant
improvements in this area would require corpus-specific additions and
tuning whose benefit would not necessarily carry over to other
corpora. In the application we are currently using the system
for---automatic extraction of subcategorisation frames, and more
generally argument structure, from large amounts of text (Briscoe \&
Carroll, 1996)---we do not need full coverage; 70--80\% appears to be
sufficient. However, further improvements in coverage will require
some automated approach to rule induction driven by parse failure.
Since our evaluations indicate that our system achieves a good level
of accuracy with little treebank data, and that 67--75\% coverage was
achieved for English quite early in the grammar refinement effort,
porting the current system to other languages should be possible with
small-to-medium-sized treebanks (around 20K words) and feasible manual
effort (of the order of 12 person-months for grammar-writing and treebanking).
This may yield a system accurate enough for some types of application, given
that the system is not restricted to returning the single highest ranked
analysis but can return the {\it n}-highest ranked for further
application-specific selection.

Although we report promising results, parse selection that is
sufficiently accurate for many practical applications will require a
more lexicalised system.  Magerman's (1995) parser is an extension of
the history-based parsing approach developed at IBM (Black \etal,
1993) in which rules are conditioned on lexical and other (essentially
arbitrary) information available in the parse history. In future work,
we intend to explore a more restricted and semantically-driven version
of this approach in which, firstly, probabilities are associated with
different subcategorisation possibilities, and secondly, alternative
predicate-argument structures derived from the grammar are ranked
probabilistically. However, the massively increased coverage obtained
here by relaxing subcategorisation constraints underlines the need to
acquire accurate and complete subcategorisation frames in a
corpus-driven fashion, before such constraints can be exploited
robustly and effectively with free text.

\section*{REFERENCES}
\newcommand{\book}[4]{\item #1 #4. {\it #2}. #3.}
\newcommand{\barticle}[7]{\item #1 #7. #2. In #5 eds. {\it #4}. #6:~#3.}
\newcommand{\bparticle}[6]{\item #1 #6. #2. In #4 eds. {\it #3}. #5.}
\newcommand{\boarticle}[5]{\item #1 #5. #2. In {\it #3}. #4.}
\newcommand{\farticle}[6]{\item #1 #6. #2. In #4 eds. {\it #3}:~#5. Forthcoming.}
\newcommand{\uarticle}[5]{\item #1 #5. #2. In #4 eds. {\it #3}. Forthcoming.}
\newcommand{\jarticle}[6]{\item #1 #6. #2. {\it #3} #4:~#5.}
\newcommand{\particle}[6]{\item #1 #6. #2. In {\it Proceedings of the
#3},~#4. #5.}
\newcommand{\lazyparticle}[5]{\item #1 #5. #2. In {\it Proceedings of the
#3}, #4.}
\newcommand{\lazyjarticle}[4]{\item #1 #4. #2. {\it #3}.}
\newcommand{\lazyfjarticle}[4]{\item #1 #4. #2. {\it #3}. Forthcoming.}

\begin{list}{}
   {\leftmargin 0pt
    \itemindent 0pt
    \itemsep 2pt plus 1pt
    \parsep 2pt plus 1pt}

\book{Alshawi, H., Carter, D., Crouch, R., Pulman, S., Rayner, M., \& Smith, A.}
{CLARE: a contextual reasoning and cooperative response framework for the Core Language Engine}
{SRI International, Cambridge, UK}
{1992} 

\particle{Billot, S. \& Lang, B.}
         {The structure of shared forests in ambiguous parsing}
         {27th Meeting of Association for Computational Linguistics}
         {Vancouver, Canada}
         {143--151}
         {1989}

\book{Black, E., Garside, R. \& Leech, G. (eds.)}
     {Statistically-driven computer grammars of English: the
IBM/ Lancaster approach}
     {Amsterdam, The Netherlands: Rodopi}
     {1993}

\lazyparticle{Brill, E.}
{Some advances in transformation-based part of speech tagging}
{12th National Conference on Artificial Intelligence (AAAI-94)}
{Seattle, WA}
{1994}
 
\barticle{Briscoe, E.}
         {Prospects for practical parsing of unrestricted text: robust
statistical parsing techniques}
         {97--120}
         {Corpus-based Research into Language}
         {Oostdijk, N \& de Haan, P.}
         {Rodopi, Amsterdam}
         {1994}

\jarticle{Briscoe, E. \& Carroll, J.}
     {Generalised probabilistic LR parsing for unification-based grammars}
     {Computational Linguistics}
     {19.1}
     {25--60}
     {1993}

\book{Briscoe, E. \& Carroll, J.}
      {Parsing (with) punctuation etc}
      {Rank Xerox Research Centre, Grenoble, MLTT-TR-007}
      {1994}

\particle{Briscoe, E. \& Carroll, J.}
{Developing and evaluating a probabilistic LR parser of part-of-speech
and punctuation labels}
{4th ACL/SIGPARSE International Workshop on Parsing Technologies}
{Prague, Czech Republic}
{48--58}
{1995}

\book{Briscoe, E. \& Carroll, J.}
{Automatic extraction of subcategorization from corpora}
{Under review}
{1996}

\particle{Briscoe, E., Grover, C., Boguraev, B. \& Carroll, J.}
         {A formalism and environment for the development of a large 
          grammar of English}  
         {10th International Joint Conference on Artificial Intelligence} 
         {Milan, Italy}
         {703--708}
         {1987}

\book{Carroll, J.}
     {Practical unification-based parsing of natural language}
     {Cambridge University, Computer Laboratory, TR-314}
     {1993}

\particle{Carroll, J.}
         {Relating complexity to practical performance in parsing with
wide-coverage unification grammars}
         {32nd Meeting of Association for Computational Linguistics}
         {Las Cruces, NM}
         {287--294}
         {1994}

\barticle{Carroll, J. \& Grover, C.}
         {The derivation of a large computational lexicon for English 
          from LDOCE} 
         {117--134}
         {Computational Lexicography for Natural Language Processing}
         {Boguraev, B. \& Briscoe, E.}
         {Longman, London}
         {1989}

\particle{Church, K.}
         {A stochastic parts program and noun phrase parser for
unrestricted text}
         {2nd Conference on Applied Natural Language Processing}
         {Austin, Texas}
         {136--143}
         {1988}

\book{Elworthy, D.}
     {Part-of-speech tagging and phrasal tagging}
     {Acquilex-II Working Paper 10, Cambridge University Computer
Laboratory (can be obtained from {\it cide@cup.cam.ac.uk})}
     {1993}
      
\lazyparticle{Elworthy, D.}
         {Does Baum-Welch re-estimation help taggers?}
         {4th Conference on Applied NLP}
         {Stuttgart, Germany}
         {1994}

\book{Garside, R., Leech, G. \& Sampson, G.}
     {Computational analysis of English}
     {Harlow, UK: Longman}
     {1987}

\particle{Grishman, R., Macleod, C. \& Meyers, A.} {Comlex syntax: building
a computational lexicon} {International Conference on Computational
Linguistics, COLING-94} {Kyoto, Japan} {268--272} {1994}

\lazyparticle{Harrison, P., Abney, S., Black, E., Flickenger, D., Gdaniec,
C., Grishman, R., Hindle, D., Ingria, B., Marcus, M., Santorini, B.
\& Strzalkowski, T.}
         {Evaluating syntax performance of parser/grammars of English}
         {Workshop on Evaluating Natural Language Processing Systems}
         {ACL}
         {1991}

\particle{Hindle, D.} 
         {Acquiring disambiguation rules from text}
         {27th Annual Meeting of the Association for Computational Linguistics}
         {Vancouver, Canada}
         {118--25}
         {1989}

\book{Jackendoff, R.}
     {X-bar syntax}
     {Cambridge, MA: MIT Press}
     {1977}

\lazyparticle{Jones, B.}
             {Can punctuation help parsing?}
             {International Conference on Computational Linguistics, COLING-94}
             {Kyoto, Japan}
             {1994}

\lazyparticle{Magerman, D.}
          {Statistical decision-tree models for parsing}
          {33rd Annual Meeting of the Association for Computational Linguistics}
          {Boston, MA}
          {1995}

\particle{de Marcken, C.}
         {Parsing the LOB corpus}
         {28th Annual Meeting of the Association for Computational Linguistics}
         {New York}
         {243--251}
         {1990}

\book{Nunberg, G.}
     {The linguistics of punctuation}
     {CSLI Lecture Notes 18, Stanford, CA}
     {1990}

\jarticle{Pereira, F. \& Warren, D.}
         {Definite clause grammars for language analysis -- a survey
          of the formalism and a comparison with augmented transition
          networks} 
         {Artificial Intelligence}
         {13.3}
         {231--278}
         {1980}

\book{Sampson, G.} {English for the computer} {Oxford, UK: Oxford University
Press} {1995}
 
\lazyparticle{Schabes, Y., Roth, M. \& Osborne, R.}
         {Parsing of the Wall Street Journal with the inside-outside
algorithm} 
         {Meeting of European Association for Computational
Linguistics}
         {Utrecht, The Netherlands}
         {1993}

\particle{Taylor, L., Grover, C. \& Briscoe, E.}
{The syntactic regularity of English noun phrases}
{4th European Meeting of the Association for Computational Linguistics}
{Manchester, UK}
{256--263}
{1989}
 
\book{Taylor, L. \& Knowles, G.}
     {Manual of information to accompany the SEC corpus:
the machine-readable corpus of spoken English}
     {University of Lancaster, UK, Ms}
     {1988}

\jarticle{Weischedel, R., Meteer, M., Schwartz, R., Ramshaw, L. \& Palmucci J.}
{Coping with ambiguity and unknown words through probabilistic models}
{Computational Linguistics}
{19(2)}
{359--382}
{1993}

\end{list}

\section*{APPENDIX}

Below is a random sample of the 250-sentence test set. The test set comprises
the Brown genre categories: ``press reportage''; ``belles lettres, biography,
memoirs''; and ``learned (mainly scientific and technical) writing''.
\begin{quote}
{\it ``Yes, your honour'', replied Bellows.\\[1mm]
This is another of the modifications of policy on Laos that the Kennedy
administration has felt compelled to make.\\[1mm]
On Monday, the Hughes concern was formally declared bankrupt after its directors
indicated they could not draw up a plan for reorganization.\\[1mm]
Ierulli will replace Desmond D. Connall who has been called to active military
service but is expected back on the job by March 31.\\[1mm]
Place kicking is largely a matter of timing, Moritz declared.\\[1mm]
Ritchie walked up to him at the magazine stand.\\[1mm]
Hector Lopez, subbing for Berra, smashed a 3-run homer off Bill Henry during
another 5-run explosion in the fourth.\\[1mm]
That's how he first won the Masters in 1958.\\[1mm]
Cooperman and Teller are accused of selling \$4,700 worth of heroin to a convicted
narcotics peddler, Otis Sears, 45, of 6934 Indiana av.\\[1mm]
However, the system is designed, ingeniously and hopefully, so that no one man
could initiate a thermonuclear war.\\[1mm]
He bent down, a black cranelike figure, and put his mouth to the ground.\\[1mm]
Those who actually get there find that it isn't spooky at all but as brilliant
as a tile in sunlight.\\[1mm]
Others look to more objective devices of order.\\[1mm]
What additional roles has the scientific understanding of the 19th and 20th
centuries played?\\[1mm]
If we look at recent art we find it preoccupied with form.\\[1mm]
Hence the beatniks sustain themselves on marijuana, jazz, free swinging poetry,
exhausting themselves in orgies of sex; some of them are driven over the borderline
of sanity and lose contact with reality.\\[1mm]
Heidenstam could never be satisfied by surface.\\[1mm]
Individual human strength is needed to pit against an inhuman condition.\\[1mm]
The pressure gradient producing the jet is due to the nature of the magnetic
field in the arc (rapid decrease of current density from cathode to the
anode).\\[1mm]
At 100 Amp the 360 cycle ripple was less than 0.5 V (peak to peak) with
a resistive load.}
\end{quote}

\end{document}